\documentclass[11pt,a4paper]{article}
\usepackage[hyperref]{tto2019}
\usepackage{times}
\usepackage{latexsym}
\usepackage{microtype}
\usepackage{graphicx}
\usepackage{booktabs} 
\usepackage[utf8]{inputenc}
\usepackage[utf8]{inputenc} 
\usepackage[T1]{fontenc}    
\usepackage{hyperref}       
\usepackage{url}            
\usepackage{booktabs}       
\usepackage{nicefrac}       
\usepackage{microtype}      
\usepackage{bm}
\usepackage{enumitem}
\usepackage[utf8]{inputenc}
\usepackage{graphicx}
\usepackage{amsmath}
\usepackage{amsfonts}       
\usepackage{amssymb}
\usepackage[english]{babel}
\usepackage{subcaption}
\usepackage[ruled,vlined,linesnumbered]{algorithm2e}
\usepackage{color, soul}

\newtheorem{definition}{Definition}
\newtheorem{assumption}{Assumption}
\usepackage{wrapfig} 
\usepackage{caption}
\usepackage{subcaption}

\usepackage{hyperref}

\usepackage{url}

\aclfinalcopy 

\title{Improving Model Robustness Using Causal Knowledge}

\author{Trent Kyono \\
  UCLA Computer Science Department \\
  Los Angeles, CA 90095\\
  \texttt{kyono@cs.ucla.edu} \\\And
  Mihaela van der Schaar \\
  UCLA Electrical Eng. Department \\
  Los Angeles, CA 90095\\
  \texttt{mihaela@ee.ucla.edu} \\}

\date{}

\begin{document}
\maketitle

\begin{abstract}

For decades, researchers in fields, such as the natural and social sciences, have been verifying causal relationships and investigating hypotheses that are now well-established or understood as truth. These  causal mechanisms are properties of the natural world, and thus are invariant conditions regardless of the collection domain or environment. We show in this paper how prior knowledge in the form of a causal graph can be utilized to guide model selection, i.e., to identify from a set of trained networks the models that are the most robust and invariant to unseen domains. Our method incorporates prior knowledge (which can be incomplete) as a Structural Causal Model (SCM) and calculates a score based on the likelihood of the SCM given the target predictions of a candidate model and the provided input variables.  We show on both publicly available and synthetic datasets that our method is able to identify more robust models in terms of generalizability to unseen out-of-distribution test examples and domains where covariates have shifted. 

\end{abstract}

\section{Introduction}


Machine learning methods, in particular neural networks, in practice are impeded by several criticisms of model assurance. Particularly, they are treated as black-box function estimators that are optimized to correctly predict for unseen data from the same distribution.  The problem becomes significantly harder if training and test data do not have the same distributions and come from various domains.  For prediction tasks, models are selected based on statistical measures such as validation accuracy or L1/L2 loss and typically have no knowledge or concern for cause and effect relationships. Because of this, they are susceptible to learning spurious associations or biases that may result in a falsely low testing error and sub-par generalizability when put into real-world operation.  These biases can be attributed to many factors, such as missing data, low sample size, or measurement error to name a few.  These issues related to domain adaptation are closely studied in the machine learning literature, and are of heightened importance when lives or costly assets are involved, such as in the healthcare industry.

Although there are many problem instances where the causal relationships between variables may be known ahead of time or can be discovered from observational data, this prior causal knowledge is seldom leveraged in predictive machine learning tasks.
We draw motivation from the notion that causality is a property of the physical world, and therefore must be invariant to any mechanism which attempts to model it. In other words, any model making a prediction based on observable variables must strive to preserve the causal relationships between the input variables and the provided prediction.  
In this work, we utilize the graphical approach from \citet{pearl2009causality} using Structural Causal Models (SCM).  
We aim to investigate and provide a new model selection criterion that leverages prior causal knowledge to improve neural network performance robustness (in terms of generalization to  unseen data) that extends beyond any traditional statistical evaluation metric.  

Our primary contribution is to provide and investigate a new selection criterion and metric that leverages prior causal knowledge in the form of a causal graph to improve neural network performance robustness. 
The main idea is to select the model whose prediction is not only statistically accurate (by for example L1/L2 loss), but also \textit{causally assured}.  
\textit{Causally assured} refers to the models whose predictions from a set of variables least violates the known causal relationships that are captured in the structure of the causal graph representing the underlying data generating process (DGP).  
We propose a proof-of-concept implementation of our approach, and show that our method is able to select the models that better generalize to unseen out-of-distribution (OOD) examples as well as causal domains where the DGP has been perturbed with noise.  
We provide experimentation results on both synthetic and several real-world examples.  
Although our method is applicable to machine learning models in general, in this work we focus on neural networks due to the inherent stochasticity of model training. Additionally, we will show how our method can further supplement and improve the performance of existing state-of-the-art causal domain adaptation algorithms  \cite{causal_transfer_jmlr, causal_domain_nips} that aim to find domain invariant features for prediction without any assumptions on the linearity of the underlying functional relationships.

\begin{figure}[!htbp]
    \centering
    \begin{subfigure}[b]{0.25\textwidth}
        \centering
        \includegraphics[width=\textwidth]{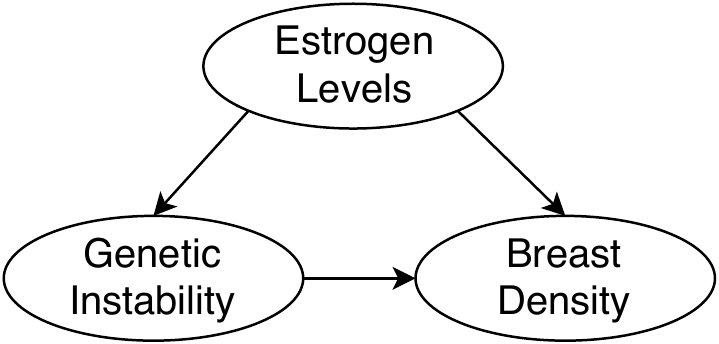}
        \caption[]%
        {}    
        \label{fig:dag1}
    \end{subfigure}
    \hfill
    \begin{subfigure}[b]{0.23\textwidth}  
        \centering 
        \includegraphics[width=\textwidth]{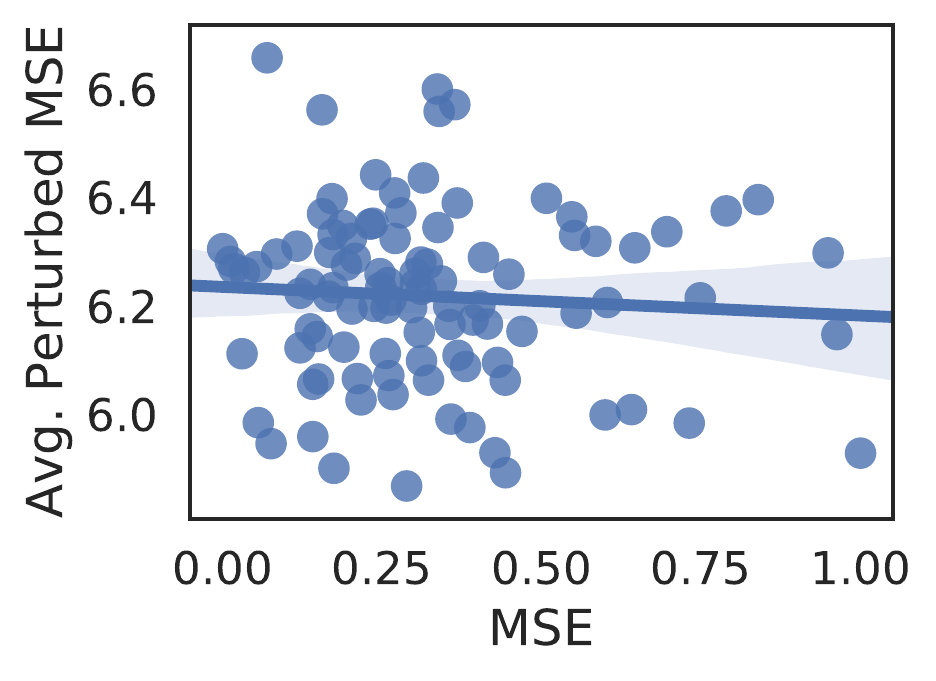}
        \caption[]%
        {}    
        \label{fig:ex1mse}
    \end{subfigure}
    \hfill
    \begin{subfigure}[b]{0.23\textwidth}  
        \centering 
        \includegraphics[width=\textwidth]{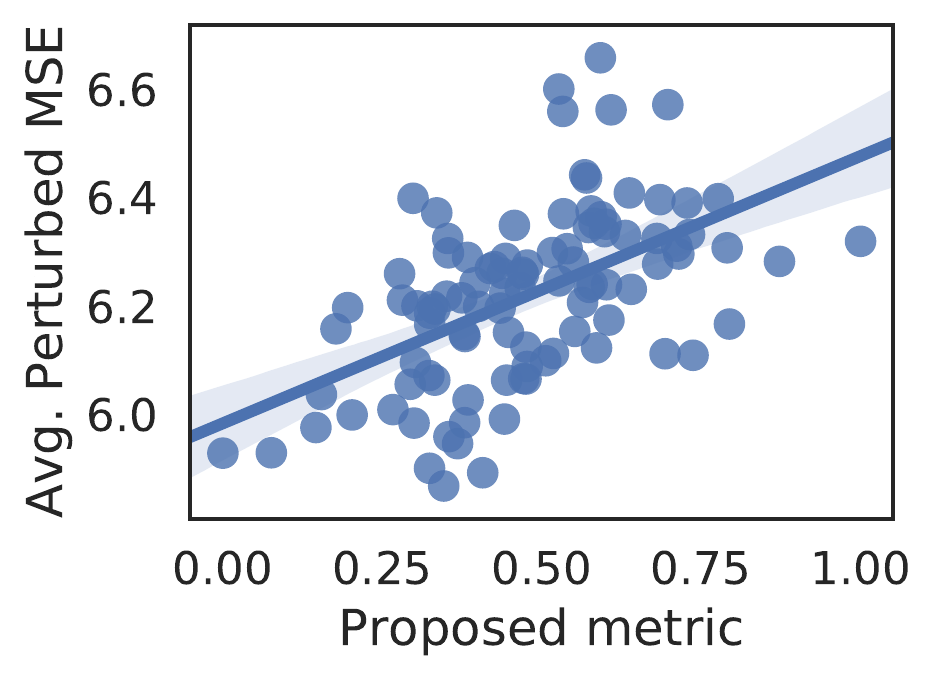}
        \caption[]%
        {}    
        \label{fig:ex1proposed}
    \end{subfigure}

    \caption[ ]
    {Motivating demonstration of our proposed metric. 100 identical deep learning models were trained to optimize MSE on the dataset generated according to the causal graph in (a). (b) shows the validation MSE versus perturbed MSE, i.e., the average error applied over the perturbed datasets. We see very little correlation between validation MSE and perturbed MSE.  In (c), our proposed metric versus the perturbed MSE shows improved coherence for predicting generalization error from our proposed metric. Each $x$-axis of (b) and (c) are normalized between 0 and 1.} 
    \label{fig:motivation}
\end{figure}

\textbf{A motivating example}  Breast density is a known risk factor for breast cancer \cite{age-breastdensity}.  Consider the causal graph shown in Fig.~\ref{fig:motivation} showing the causal relationships between estrogen levels, genetic instability, and breast density \cite{dag1}.  Here, we are interested in predicting a patient's breast density from their estrogen levels and measured genetic instability.  
This causal structure is a property of the physical world and is invariant to distribution noise or shifts. 
Therefore, we can use this structure to select the most robust model in terms of generalizability.  We implemented a system of causal functions with functional relationships between adjacent variables and i.i.d. Gaussian noise applied to each to generate a synthetic dataset consistent with the graph structure in Fig.~\ref{fig:dag1}.  We trained 100 deep learning models to predict breast density from a patient's estrogen levels and genetic instability.  Each of the models were trained with identical hyperparameters (layers, initialization, etc.) and training regimes on the synthetic dataset generated by our system of causal functions with Gaussian noise at zero mean and unit variance. 
To simulate these various perturbed distributions, we generated 9 modified test sets (using our same causal system) each with a unique set of combinations of means and variances from the range $(0,2)$ and $(1, 3)$, respectively.  
In Fig.~\ref{fig:ex1mse}, for each of our 100 models we plot the validation MSE versus average perturbed MSE.  It is evident that the validation MSE is not a good method for selecting the model that will generalize well.  However, in Figure~\ref{fig:ex1proposed} for the same 100 models, we show significantly improved coherence between our proposed metric and the average perturbed MSE (a measure of generalization robustness). 

\section{Related works}

There exists a plethora of research addressing domain adaptation, a sub-task in the field of transfer learning. 
For a general overview we refer to \citet{pan_and_yang_survey}.   
SCMs have been studied for domain adaptation by leveraging the invariance of the causal graph that describes the underlying DGP for years, and has origins as early as \citet{spirtes2000} or perhaps even earlier.  
Some notable works include  \citet{causal_invariance1}, who demonstrate how causal knowledge can facilitate and dictate machine learning approaches. \citet{Bareinboim:2012} provide a theory for identifiability under transportability assuming that a causal graph and the intervention targets are known. 
\citet{causal_domainICML2013, causaldomain_aaai} assume perfect interventions with known targets as well, but expand the methods to more than SCMs. 
\citet{causal_domain_nips} and \citet{causal_transfer_jmlr} attempt to identify some subset of covariates that will lead to the most domain transferable predictions for linear Gaussian models.  
\citet{causal_transfer_jmlr} assumes that if there is invariance of a conditional distribution in various source domains, then the conditional distribution will be the same in the target domain.  They operate under the assumption that the intervention and test domain will be known ahead of time and select invariant predictive features accordingly. 
\citet{causal_domain_nips} does not make such assumption, and addresses the setting in which both the causal graph and intervention type and targets may be partially unknown.  
However, \citet{causal_domain_nips} points out that such an invariant set may not exist or their algorithm may not converge on such a set.   
Because of this, rather than reduce or search for a subset of invariant features, we approach the problem of domain generalization from a different direction. Rather, our method is a selective approach that can be added to any machine learning method to improve domain generalization.  We will show in the experimental section how our method can be used in conjunction with causal domain adaptation feature selection methods of \citet{causal_domain_nips} or  \citet{causal_transfer_jmlr} to improve generalization to perturbed domains.  An overview of related works are provided in Table~\ref{table:related-works}.

Although our methodology generalizes to any ``black-box'' machine learning model, our focus here is on neural networks due to its stochastic training properties (we show experiments on a handful of other machine learning models in Appendix~B).  Training a neural network requires minimizing a high-dimensional non-convex loss function with no guarantees on optimizing to the same minima over two consecutive runs. Choices for optimization, such as learning rate, stopping condition, initializations, etc., and many common techniques used in training, such as stochastic gradient descent (SGD) or dropout layers, further obfuscate reproducibility \cite{deep_learning_generalization,deep_learning_2}. Because of this, it is important to assess the neural network performance by more than just a statistical loss function, which may not differentiate between closely spaced minima.  Many techniques have been proposed for deep learning generalization (for a summary see \citet{generalization}), but as far as we know, our approach to this problem is unique in that we discipline our model selection  based on the preservation of the causal structure of the machine learning model predictions and input variables relative to our prior causal or human knowledge of the DGP.  


\begin{table}[!htbp]
\caption{Overview of related causal domain transfer methods.  ML (general machine learning) is checked if the method applies to general machine learning (rather than just SCMs). PG (partial DAGs) is checked if the method applies to methods with partial graphs (incomplete causal DAGs).  IA (intervention agnostic) is checked when the method is agnostic to the intervention/perturbation location in the DAG. NL (linearity assumption) is checked if the method does not make any assumptions on linearity of underlying functional connections. MS (model selection) is checked when the method can be used for model selection.}
\label{table:related-works}
\vskip 0.15in
\begin{center}
\begin{small}
\begin{sc}
\setlength\tabcolsep{3pt}
\begin{tabular}{lccccc}
\toprule
Method & ML &  PG & IA & NL & MS  \\
\midrule
\citet{Bareinboim:2012}         &           &                   &          & $\surd$ &       \\
\citet{causal_domainICML2013}   &  $\surd$         &           &           & $\surd$ &       \\
\citet{causaldomain_aaai}       &  $\surd$         &          &            & $\surd$ &      \\
\citet{causal_transfer_jmlr}    & $\surd$   &   $\surd$       &             & &       \\
\citet{causal_domain_nips}      & $\surd$   &   $\surd$        & $\surd$   & &       \\
Proposed                        & $\surd$   &   $\surd$        & $\surd$   & $\surd$ & $\surd$\\
\bottomrule
\end{tabular}
\end{sc}
\end{small}
\end{center}
\vskip -0.1in
\end{table}

\section{Problem Formulation}

\subsection{Preliminary notation and assumptions}

In this work we base our notation on the causal framework of \citet{pearl2009causality}.  A causal structure of a set of variables $V$ is a directed acyclic graph (DAG) in which each vertex corresponds to a distinct element in $V$, and each link represents direct functional relationships between the corresponding variables.  An SCM is a pair  $M = \langle G, \Theta_G \rangle$ consisting of a causal structure, $G$, and a set of parameters $\Theta_G$ compatible with $G$. The parameters, $\Theta_G$, assign a function $x_i = f_i(pa_i, u_i)$ to each $X_i \in V$ and a probability measure $P(U_i)$ to each $U_i$, where $PA_i$ represent the parents (direct causes) of $X_i$ in $G$ and where each $U_i$ is some i.i.d disturbance according to $P(U_i)$.  

We operate under several assumptions.  Our primary assumption, which we will refer to as \textit{causal invariance} is:

\begin{assumption}[Causal invariance]\label{as:ca}
Let $G$ be a causal DAG representing variables $V$, $P(V)$ be the corresponding distribution on $V$, $E$ be  a set of environments or domains, and $I(\langle P(V), e \rangle)$ denote the set of all conditional independence relationships embodied in $P(V)$ for a domain $e \in E$, then $(\forall e_i, e_j \in E:  I(\langle P(V), e_i \rangle) = I(\langle P(V), e_j\rangle)$.
  \end{assumption}
Assumption~\ref{as:ca} states that the functional relationships of variables to their direct causes are invariant across any set of environments.  Because of this, the conditional independence relationships and DAG structure are invariant across domains as well.  Similar assumptions have been made in other works \cite{causal_invariance1,causal_invariance,causal_invariance2,causal_domain_nips, causal_transfer_jmlr}.  
Because we perform causal discovery in this work, we assume the standard causal discovery \cite{pearl2009causality,gfci} assumptions that include the following: (i) the DGP can be represented by an SCM, where the data has a functional relationship among the measured variables with finite Gaussian noise, (ii) there are no cyclical dependencies or feedback loops  (causal structures adhere to a DAG structure), (iii)  all variables are observed (no unobserved hidden confounders), and (iv) the Markov assumption and faithfullness holds.  The causal Markov assumption states that a variable is independent of its non-descendants, given its parents, and faithfulness states that all the independence relationships among the measured variables are implied by the causal Markov assumption.

\begin{figure*}[!htbp]
\centering
    \includegraphics[width=0.8\linewidth]{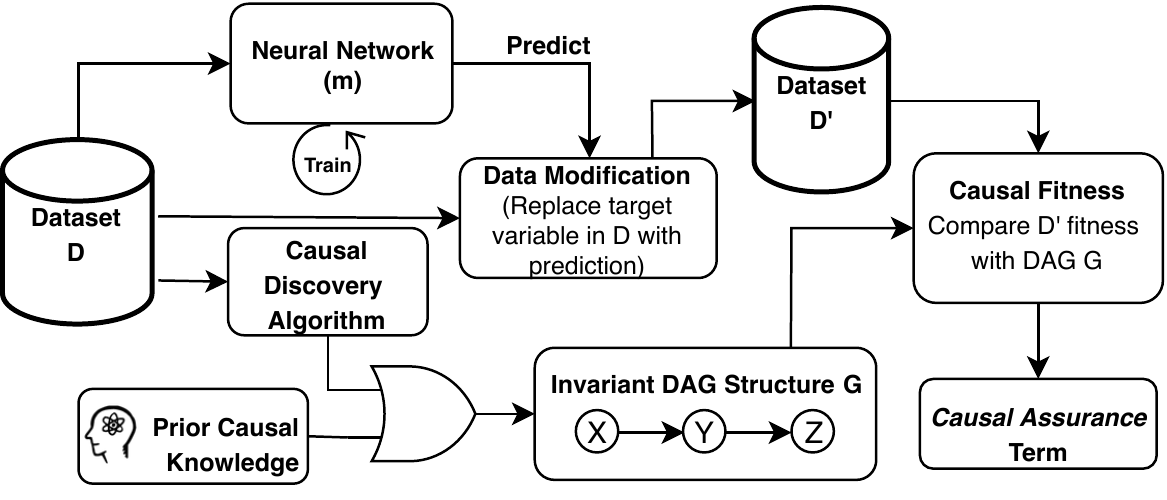}
\caption{Schematic of \textit{causal assurance} term.  Given a dataset $D$, prior knowledge or a causal discovery algorithm can be used to arrive at the invariant DAG structure $G$.  Models $m$ are trained on $D$ to predict a target variable in $D$.  Replacing the predictions of the target variable by $m$ in $D$ we generate a new dataset $D'$.  The \textit{causal assurance} term can be calculated by comparing the fitness of $D'$ to $G$ by metric scores such as the log-likelihood or BIC score.}  \label{fig:schematic}
\end{figure*}

\subsection{Causal assurance criterion}

In this section, we discuss our causal assurance criterion.  We reiterate our \textit{causal invariance} assumption that the functional relationships of variables to their direct causes are invariant across any set of environments, and therefore can be represented by the same graphical structure in any domain.  Based on our assumption of \textit{causal invariance}, we provide a general definition for \textit{causal assurance} as follows:

\begin{definition}[Causal assurance]
\label{def:causal_assurance}
Let $V$ be a set of variables, let $G$ be a DAG for the SCM of $V$, and let $\widehat{V}_i$ be the prediction of a machine learning model $m \in M$ for any of the variables $V_i \in V$. We say that $m$ preserves causality (\textit{causally assured})  if and only if there exists a discoverable DAG $\widehat{G}$ in the Markov equivalence classes of $(V - V_i) \cup \widehat{V}_i$ where  $G$ and $\widehat{G}$ have the same DAG structure.
\end{definition}

 Def.~\ref{def:causal_assurance} implies that we prefer models whose predictions, $\widehat{V}_i$, preserve the conditional independence relationships of $V$, such that our assumption of \textit{causal invariance} is not violated.  Succinctly, the causal DAG structure for the variables $(V - V_i) \cup  \widehat{V}_i$ (in any domain) is the same as the true causal DAG structure $G$.  A score for a \textit{causally assured} model would take into consideration both the model's predictive performance as well as a distance metric for how causally compliant said model's predictions are to the causal DAG representing the underlying DGP.  Therefore, we define a metric for \textit{causal assurance} as follows:


\begin{definition}[Causal assurance metric (CAM)]
\label{def:causal_assurance_met}
Let $m$ be a trained model, $D$ be a dataset, and $G$ be a causal DAG for the variables in $D$.  Our CAM is provided by $r$, which is defined as:
\begin{equation}\label{eq:metric}
    r(m, D, G) = \lambda f(m, D, G) + h(m, D)\textit{,}
\end{equation}
where $f$ is a scoring function that measures the fitness of the causal DAG $G$ to the dataset $D'$ containing the variables $(V - V_i) \cup \widehat{V}_i$, and $h$ is a function that returns the error, such as MSE, of $m$'s predictions on $D$.

\end{definition}

We will refer to the output of $f$ as our \textit{causal assurance} term (see Fig.~\ref{fig:schematic}) and explain how to calculate in the next subsection.  For metrics that we wish to maximize, such as accuracy, subtract $h$ instead.  $\lambda$ is a tuning factor between our \textit{casual assurance} term and machine learning model performance. In this work, we define $\lambda$ in terms of the expected performance of $f$ and $h$ and the uncertainty associated with the underlying DAG.  Specifically we have defined $\lambda = \mathbb{E}(Y_f)/\gamma \mathbb{E}(Y_h)$, where $Y_f$ is the test performance of the models selected using $f$, $Y_h$ is the test performance of models selected using $h$, and $\gamma$ is the number of potential other DAGs in the Markov Equivalence Class (MEC) of $G$.  This will penalize our score based on the uncertainty or probability that we know the true DAG $G$.  For example, if we are absolutely certain through experimentation (such as randomized trials) that $G$ is the correct DAG then $\gamma = 1$.  However, in many cases their may be several other potential DAGs in the MEC of $G$ that explain our observations.  If this is the case, then the value of $\lambda$ is decreased by a factor of $\gamma$ accordingly.  If choosing between different classes or types of machine learning models, an additional term could be added to Eq.~\ref{eq:metric} to represent machine learning model complexity.  In this work we only consider selection between identical neural network architectures.  The selection of the most \textit{causally assured} model in $M$ can be found by finding the model with the lowest score using our function $r$, such that: 

\begin{equation}\label{eq:min}
    \min_{m \in M}  r(m, D, G)
\end{equation}

 \citet{causal_transfer_jmlr} show that for predictive models when location of the interventional perturbation is not known, the invariant set of predictors is the causal parents of the target variable.  This stems from the fact that an intervention of a child node will not propagate in the anti-causal direction to its parents \cite{causal_invariance1}.  Because we make no assumption to the location of any perturbation in $G$, throughout the remainder of this manuscript, assume that the outgoing edges of the target $V_i$ have been removed in $G$ (rendering $V_i$ conditionally independent of any child) when calculating our \textit{causal assurance} metric.

\begin{algorithm}[!ht]
\DontPrintSemicolon
\SetAlgoLined
\SetKwInOut{Input}{Input}\SetKwInOut{Output}{Output}
\Input{A selection dataset $D$ with set of variables $V$, target variable $V_i \in V$, a set of untrained models $M$, a method for measuring data fitness to DAG structure $l$, an assumed \textit{causally invariant} DAG structure $G$.}
\Output{The most \textit{causally assured} model $M$}
\BlankLine

Divide $D$ into three disjoint sets $D_{train}$, $D_{val}$, and $D_{sel}$ for model training, validation, and selection, respectively. 

Train each model $m \in M$ on $D_{train}$ until performance of $m$ on $D_{val}$ converges (stops improving).

\For{$m \in M$}{
    
 Use $m$ to predict  $\widehat{V}_i$ given variables $V - V_i$ in $D_{sel}$.
 
 Generate dataset $D'_{sel}$, where $D_{sel}' \leftarrow (V - V_i) \cup \widehat{V}_i$.
 
 Calculate $f(m, D_{sel}, G)$ using $l$ on $G$ and $D_{sel}'$, and calculate $h(m,D_{sel})$ from $MSE(V_i,\widehat{V}_i)$.
 
 Store the causal assurance term $f(m, D_{sel}, G)$ and the predictive error $h(m,D_{sel})$.
}
Return $m \in M$ with lowest $\lambda f(m, D_{sel}, G) + h(m,D_{sel})$
\caption{Causal assurance selection} \label{alg:causal_assurance}
\end{algorithm}

\subsection{Model scoring and selection}

In score-based causal discovery, the Bayesian Information Criterion (BIC) is a common score that is used to discover the completed partially directed acyclic graph (CPDAG), representing all DAGs in the MEC, from observational data.  Under the Markov and faithfullness assumptions, every conditional independence in the MEC of $G$ is also in $D$. The BIC score is defined as:

\begin{equation}
    BIC(G|D) = -LL(G|D) + \left(\frac{\log_2N }{2}\right)||G||,
\end{equation}
where $N$ is the data set size and $||G||$ is the dimensionality of $G$.  For our function $f$ in Eq.~\ref{eq:metric}, we use the BIC score. However, since $N$ and $||G||$ are held constant in our proposed method our function $f \propto -LL(G|D)$.  To find the $LL(G|D)$ we use the following decomposition:

\begin{equation}
    LL(G|D) = -N \sum_{X_iPA_i}H_D(X_i|PA_i)\textit{,}
\end{equation}
where $N$ is the dataset size, $PA_i$ are the parent nodes of $X_i$ in $G$, and $H$ is the conditional entropy function which is given by \citet{darwiche} for discrete variables and by \citet{mixed-variables} for continuous or mixed variables. 

The complete causal DAG of the variables in $D$ may not be known ahead of time.  In some cases, only a few of the causal relationships may be known a priori.
If the complete DAG structure is not known ahead of time, a causal discovery algorithm can be used to determine a CPDAG.  The CPDAG represents a set of Markov equivalent graphs, from which a human expert can select the DAG, $G$, that is most copacetic with causal beliefs.  In either case, the resulting causal DAG will be assumed to be true and  \textit{causally invariant}, which we can use as a reference in our CAM.

We present an Algorithm~\ref{alg:causal_assurance} for selecting the most \textit{causally assured} model according to Eq.~\ref{eq:min}.  Algorithm~\ref{alg:causal_assurance} assumes that a causal DAG $G$ is provided either by prior causal knowledge, by causal discovery, or by some combination of the two.  Algorithm~\ref{alg:causal_assurance} takes a dataset $D$, with a set of variables $V$ and a target variable $V_i \in V$, and partitions $D$ into three disjoint sets for training $D_{train}$, model validation $D_{val}$, and model selection $D_{sel}$.   A separate selection set $D_{sel}$ is used to discourage any bias from model validation into model selection.  The candidate models $M$ are trained on $D_{train}$ until they converge on $D_{val}$, i.e., the loss calculated on $D_{val}$ stops improving.  For each model $m \in M$, a new dataset $D'_{sel}$ is created from $(V - V_i) \cup \widehat{V}_i$, where $\widehat{V}_i$ is the predictions of $m$.  A measure of data fitness to DAG structure $l$ is used on the dataset $D'_{sel}$ and $G$ to generate the causal assurance term $f(m, D_{sel}, G)$, and $h(m,D_{sel})$ can be calculated from the predictive error of $m$, such as MSE.  Optionally, if the selection set is large enough, an average value for $f(m, D_{sel}, G)$ can be calculated by dividing $D'_{sel}$ into $k$-folds.    The output of Algorithm~\ref{alg:causal_assurance} is the model with the lowest CAM.

    \begin{figure}[!htbp]
        \centering
        \begin{subfigure}[b]{0.23\textwidth}
            \centering
            \includegraphics[width=\textwidth]{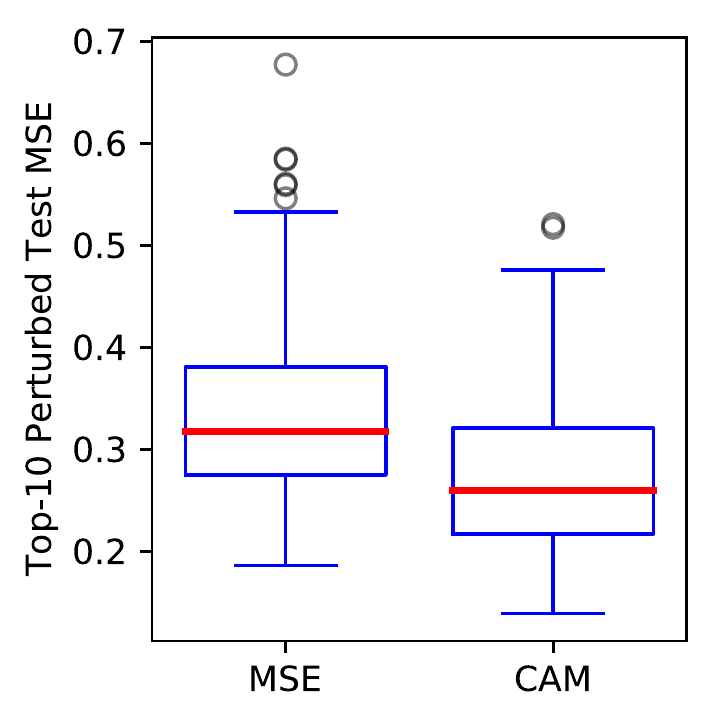}
            \caption[]%
            {MSE-10 vs CAM-10}    
            \label{fig:single_intervention}
        \end{subfigure}
        \hfill
        \begin{subfigure}[b]{0.23\textwidth}
            \centering
            \includegraphics[width=\textwidth]{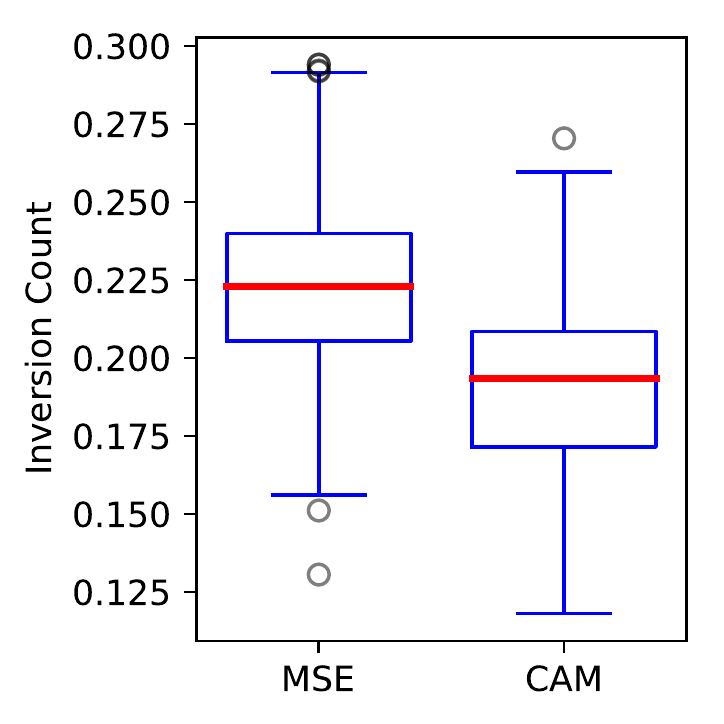}
            \caption[]%
            {Inversion Count}    
            \label{fig:multiple_intervention}
        \end{subfigure}

        \caption[ ]
        {Performance of proposed metric (CAM) on synthetic experiments of various vertex cardinalities.  (a) We show our proposed method in terms of selecting the best models using MSE-10 and CAM-10. CAM-10 is the predictive performance on our perturbed test set of the top 10\% of the models selected by our CAM.  Similarly, MSE-10 is the performance of the top 10\% of the models selected by MSE. (b) We use an inversion count to show our proposed method in terms of ranking models by testing performance.} 
        \label{fig:synthetic}
    \end{figure}

For conciseness and correctness of problem exposition, we present our method in terms of model selection.  However, it is important to note that although we present our methods in terms of model selection on a selection set $D_{sel}$, our method can be trivially applied to the model validation stage as done in traditional machine learning validation (such as k-fold cross-validation).   For machine learning models that do not have the option of early stopping or terminating at a checkpoint during training, $D_{val}$ can be omitted (step 2 in Algorithm~\ref{alg:causal_assurance}), and $D_{sel}$ used as a validation set.  With this in mind, our method can be used to guide hyper-parameter selection if desired.

\section{Experiments}

Experiments were performed on both synthetic and publicly available real-world Kaggle datasets.  For the synthetic data experiments, the true causal graph was first established to generate each dataset.  Conversely, for the real-world datasets the true causal graph was not fully known, and prior or ``common sense'' knowledge was leveraged to discover the causal graph.  To maintain the scope of this paper, we focus on regression predictions only (for classification see Appendix~B), and therefore use MSE as our function $h$ in Eq.~\ref{eq:metric}.

\subsection{Evaluation metrics}

We evaluate our models based on the following performance metrics.  For each of our experiments we evaluated and ranked models according to their performance on our selection set, by either MSE or our CAM.  The first metric we propose examines the top $10\%$ of models selected by calculating the MSE loss or our CAM on the selection set to show the generalization improvement in terms of MSE on test sets.  We refer to selecting the best 10\% of the trained models by their performance on the selection dataset in terms of MSE or CAM as MSE-10 or CAM-10, respectively.  The second metric we propose examines the ability to rank the entire set of trained models (rather than selecting the top $10\%$) by either MSE or CAM in terms of test performance.  Here we use a list inversion count (IC), which measures the number of element-wise inversions required for sorting a list.  IC can be thought of as a metric of how sorted a list is, where a perfectly sorted list has an IC of 0.  The asymptotic maximum number of inversions for a list of $n$ elements is $n(n-1)/2$, which we use to normalize our IC metric.

\subsection{Robustness with complete knowledge}

We first perform experiments on synthetic data to demonstrate our method for improving generalization performance when the truth causal structure is known. For each of the simulations, we generated a random DAG, $G$, with $n$ vertices and $e \sim \mathcal{U}(n, n(n-1) / 2)$ (up to the asymptotic maximum number of edges in a DAG) edges between them.  Using the structure of $G$ we synthesized two datasets $D_1$ and $D_2$ with functional relationships (randomly chosen as linear or non-linear exponential functions) between variables with directed edges between them in $G$ and applied Gaussian noise to each. We will refer to this underlying data generation mechanism  as the DGP.  We enumerated all nodes in $G$ randomly. To prevent the magnitude of the leave nodes from becoming overwhelmingly large relative to the root nodes, each node was a function of its parents values that were either subtracted or added for even or odd enumerated parents, respectively.  $D_1$ was generated by sampling 10,000 data points with a Gaussian noise having a mean of 0 and variance of 1, and was randomly partitioned into a model training, validation, and selection set of 60\%, 20\%, and 20\%, respectively.  The validation set was used to terminate model training, and the selection set was reserved for model selection and ranking.  $D_2$ was reserved as a test set to evaluate model generalization performance.  The input features of $D_1$ were min-max normalized between 0 and 1, and the min and max values for each feature were saved for scaling $D_2$ accordingly.

For each DAG $G$, we randomly selected a target variable from $G$ that was connected to at least one other variable in $G$.  We then trained 100 deep neural networks with identical architectures on our training set to predict our target variable.  Each model had two hidden layers with the number of neurons equal to the number of input features.  Both layers were initialized with Glorot uniform weights and followed by a dropout mask of 0.2 after each.  The activation function for the hidden layers and output layer was ReLU and linear, respectively.  Models were trained using SGD for 20 epochs with a learning rate of $10^{-4}$ with a momentum of 0.9, and stopped and saved when MSE validation loss converged and failed to improve on our validation set.  
We evaluated and ranked each of the 100 models by their performance on our selection set, by either MSE or our CAM, and repeated this 100 times for each dataset. Although we do not see any practical limitation regarding the number of nodes that our method will work for, we capped our simulation experiments at 64 nodes due to the computational time complexity involved in random DAG generation which required finding the maximum number of possible edges $e \sim \mathcal{U}(n, n(n-1) / 2)$. We performed experiments for DAGs having $n \in \{4,8,16,32,64\}$ vertices.   

We performed experiments on test sets of 2000 samples.  We created our test datasets $D_2$ with at least one of the variables in $G$ randomly perturbed using our DGP with mean of 1 and a variance of 2 (rather than mean of 0 and variance of 1 as used in $D_1$ for training, validation, and selection).  The noise terms in all the remaining variables stayed the same as in $D_1$.  Fig.~\ref{fig:synthetic} shows that CAM is able to better select and rank models models in terms of perturbed test mse and inversion count, respectively.

    \begin{figure}[!ht]
        \centering
        \begin{subfigure}[b]{0.23\textwidth}
            \centering
            \includegraphics[width=\textwidth]{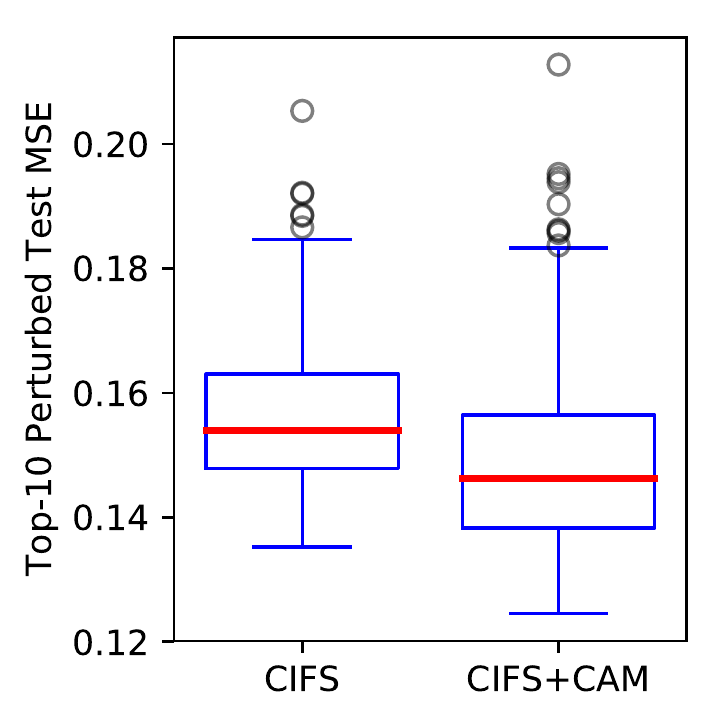}
            \caption[]%
            {MSE-10 vs CAM-10}    
            \label{fig:CIFS10}
        \end{subfigure}
        \hfill
        \begin{subfigure}[b]{0.23\textwidth}
            \centering
            \includegraphics[width=\textwidth]{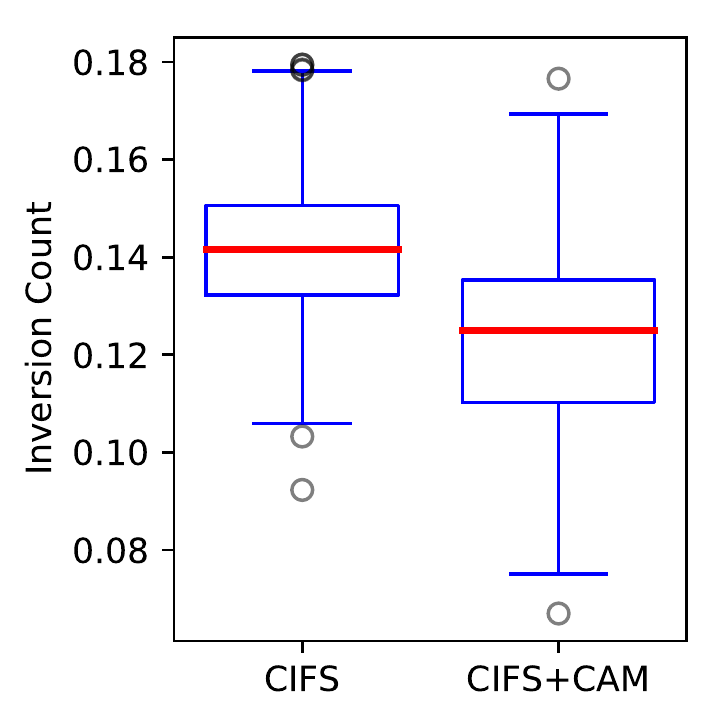}
            \caption[]%
            {Inversion Count}    
            \label{fig:CIFSIC}
        \end{subfigure}

        \caption[ ]
        {Performance of proposed metric (CAM) on synthetic experiments of various vertex cardinalities.  (a) We show our proposed method in terms of selecting the best models using MSE-10 and CAM-10. CAM-10 is the predictive performance on our perturbed test set of the top 10\% of the models selected by our CAM.  Similarly, MSE-10 is the performance of the top 10\% of the models selected by MSE. (b) We use an inversion count to show our proposed method in terms of ranking models by testing performance.} 
        \label{fig:CIFS}
    \end{figure}

\paragraph{Going beyond existing feature selection algorithms}
To show how our method can be used to improve the state-of-the-art we use the invariant feature selection methods for causal domain adaptation presented in \citet{causal_domain_nips} which is a more generalized approach than \citet{causal_transfer_jmlr}.  We used the same experimental setup mentioned in the previous subsection, but instead of using all of the input features, we use the input features identified by \citet{causal_domain_nips} that will result in the most domain transferable predictions. We will refer to this feature selection approach as CIFS (for causally invariant feature selection). In the identical synthetic setting used in the prior experiment, we apply there CIFS method to select the most invariant causal features to use as input features and we apply our selection method on 100 trained CIFS models.  We will refer to selection of the best CIFS models using our CAM as CIFS+CAM.  Fig.~\ref{fig:CIFS10} shows that our method (CIFS+CAM) is able to improve on CIFS in terms of both inversion count and top 10\% model selection.

\begin{figure}[!t]
        \begin{subfigure}[b]{0.23\textwidth}
            \centering
            \includegraphics[width=\textwidth]{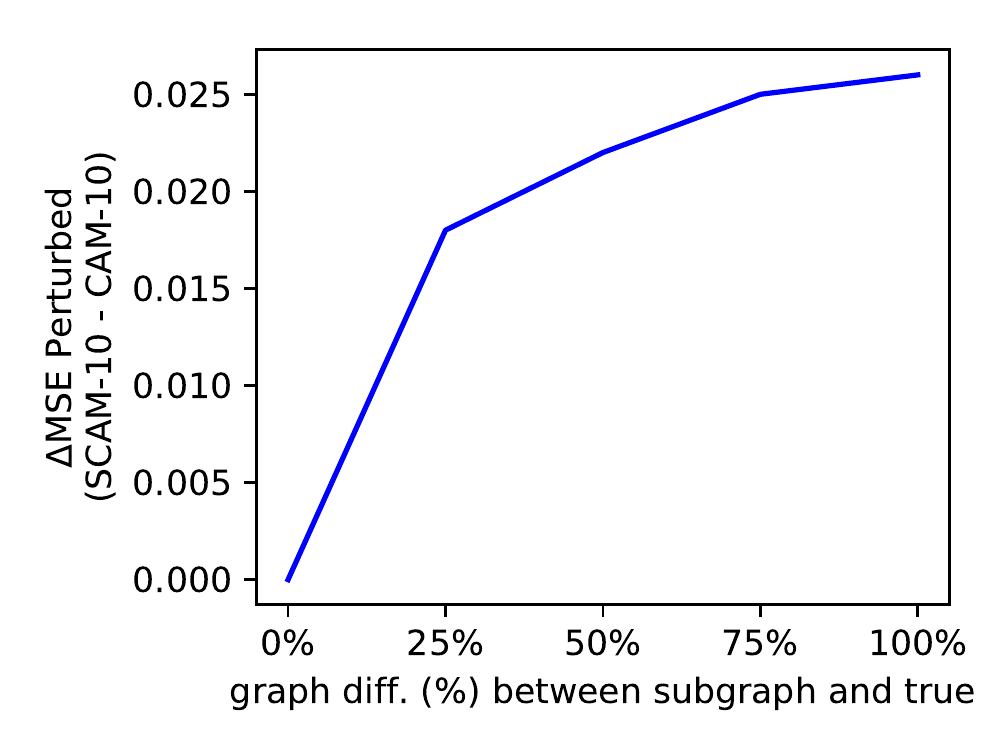}
            \caption[]%
            {Subgraph Results}    
            \label{fig:subgraph}
        \end{subfigure}
       \hfill
    \begin{subfigure}[b]{0.23\textwidth}
            \centering
            \includegraphics[width=\textwidth]{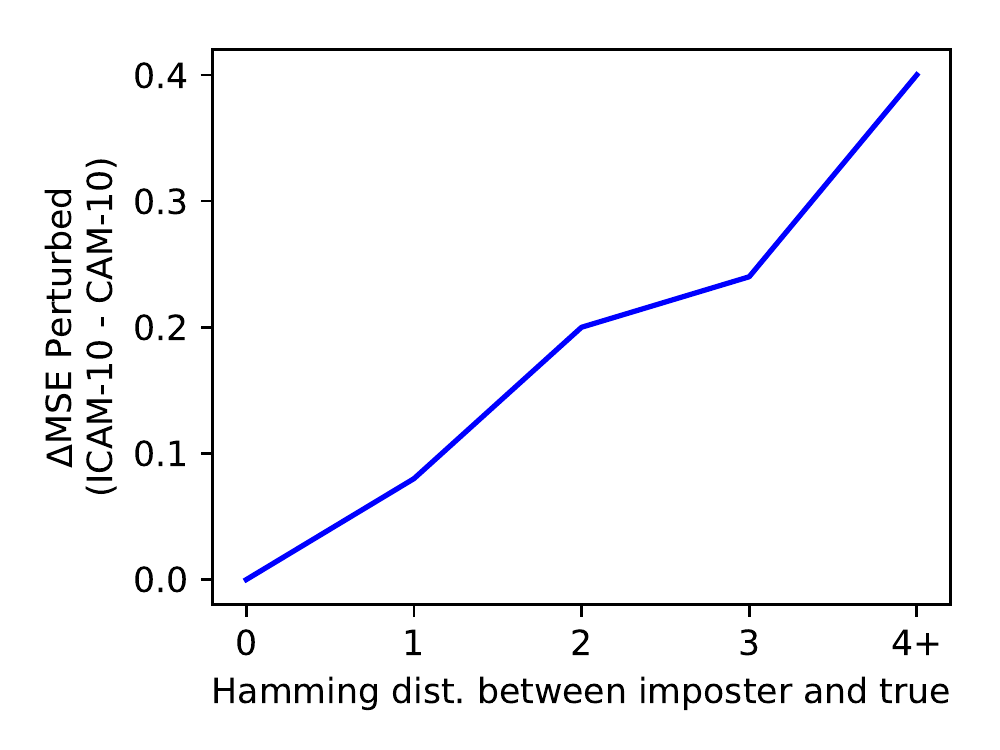}
            \caption[]%
            {Imposter Results}    
            \label{fig:imposter_sub}
        \end{subfigure}
    \vspace{-10pt}
  \caption{(a) Graph of the difference in perturbed MSE ($\Delta$MSE) of models selected using a subgraph of the truth DAG (SCAM-10) and the truth DAG (CAM-10) vs the percentage difference in terms of edges between the two graphs. (b) Graph of the difference in perturbed MSE ($\Delta$MSE) of models selected using an imposter DAG (ICAM-10) and the truth DAG (CAM-10) vs the Hamming distance between the true causal graph and the imposter graph with edges flipped or added.  ICAM-10 is the predictive performance on our perturbed test set of the top 10\% using our CAM, but with an incorrect causal DAG. }
  \label{fig:imposter}
\end{figure}

\paragraph{Sensitivity analysis on subgraphs}
We perform a sensitivity analysis on subgraphs of the true underlying causal DAG.  We use the same synthetic experimental setup, but we randomly take subsets of the underlying causal truth casual DAG.  We refer to the top-10 performance of the subset DAGs as SCAM-10.  Fig.~\ref{fig:subgraph} plots the average difference in perturbed MSE ($\Delta$MSE) of models selected using a subgraph of the truth DAG (SCAM-10) and the truth DAG (CAM-10) versus the percentage difference in terms of edges between the two graphs. At 100\% of the edges missing our CAM metric does not have any impact on the outcome and therefore is equivalent to MSE selection. Fig.~\ref{fig:subgraph} shows that even with partial knowledge our method still is able to improve selection.  Note that in practice, causal discovery should be used in addition to any subgraphs (of prior knowledge) to include the maximum number of edges possible.

\paragraph{Robustness under a DAG imposter} 
To highlight the effectiveness of causal knowledge in our proposed method, we showcase the performance degradation using our methodology when the prior knowledge may be incorrect. In the prior experiment we focused on when the prior knowledge may be correct but incomplete (in terms of missing edges).  However in this experiment we used an identical synthetic experimental setting as above, except we generate a second DAG (an imposter) with the same graphical structure but with edges into the target variable that are either added or reversed.  We generated 100 random DAGs with vertex cardinalities between 4 and 64 vertices, and use a simple Hamming distance metric to measure the similarity of the truth DAG from the imposter DAG. Fig.~\ref{fig:imposter_sub} shows the results of the experiment, where ICAM-10 selects the top 10\% of the models using our CAM but calculated with the imposter DAG instead.  From this experiment, the farther (in terms of Hamming distance) our imposter DAG is from the truth DAG, the worse our metric is at identifying robust models for generalization.  Although it is reassuring that providing the wrong DAG information does in fact result in poorer robustness (perturbed MSE), it stresses the importance of knowing the truth causal structure in our methodology.

\subsection{Out-of-distribution robustness on real data} \label{sec:real_data}
In practice, often times the complete underlying causal DAG is unknown.  In this set of experiments we explore using incomplete prior knowledge on four publicly available Kaggle datasets. The Kaggle datasets include the Abalone \cite{kaggle:abalone}, Bike Sharing in Washington D.C. \cite{kaggle:bikeshare}, Student Performance in Exams \cite{kaggle:student-performance}, and Open Powerlifting \cite{kaggle:powerlifting} datasets.  The Abalone dataset and Bike Sharing were used for causal applications in \citet{abalone_causality} and \citet{anchor_regression}, respectively. Additional dataset details and respective causal discovery steps are provided in Appendix~A. 

For each dataset we used either prior knowledge or causal discovery algorithms to determine the full causal graph.
If the complete causal graph wasn't known, to learn the remaining causal connections from the data we used the Fast Greedy Equivalence Search (FGES) algorithm by \citet{fges} on the entire dataset using the \texttt{Tetrad} software package \cite{tetrad}.  
\texttt{Tetrad} allows prior knowledge to be specified in terms of required edges that must exist, forbidden edges that will never exist, and temporal restrictions (variables that must precede other variables). 
Using our prior knowledge, we used the FGES algorithm in \texttt{Tetrad} to discover the causal DAGs (see Appendix~A) for each of the public datasets.  
Only the directed edges that were output in the CPDAG by FGES were considered as known edges in the causal graphs.
The \texttt{Tetrad} software package automatically handles continuous, discrete, and mixed connections, i.e. edges between discrete and continuous variables.  If not using \texttt{Tetrad} for mixed variables, the method from \citet{mixed-variables} can be used.

\begin{figure}[!htbp]
    \centering
        \includegraphics[width=0.35\textwidth]{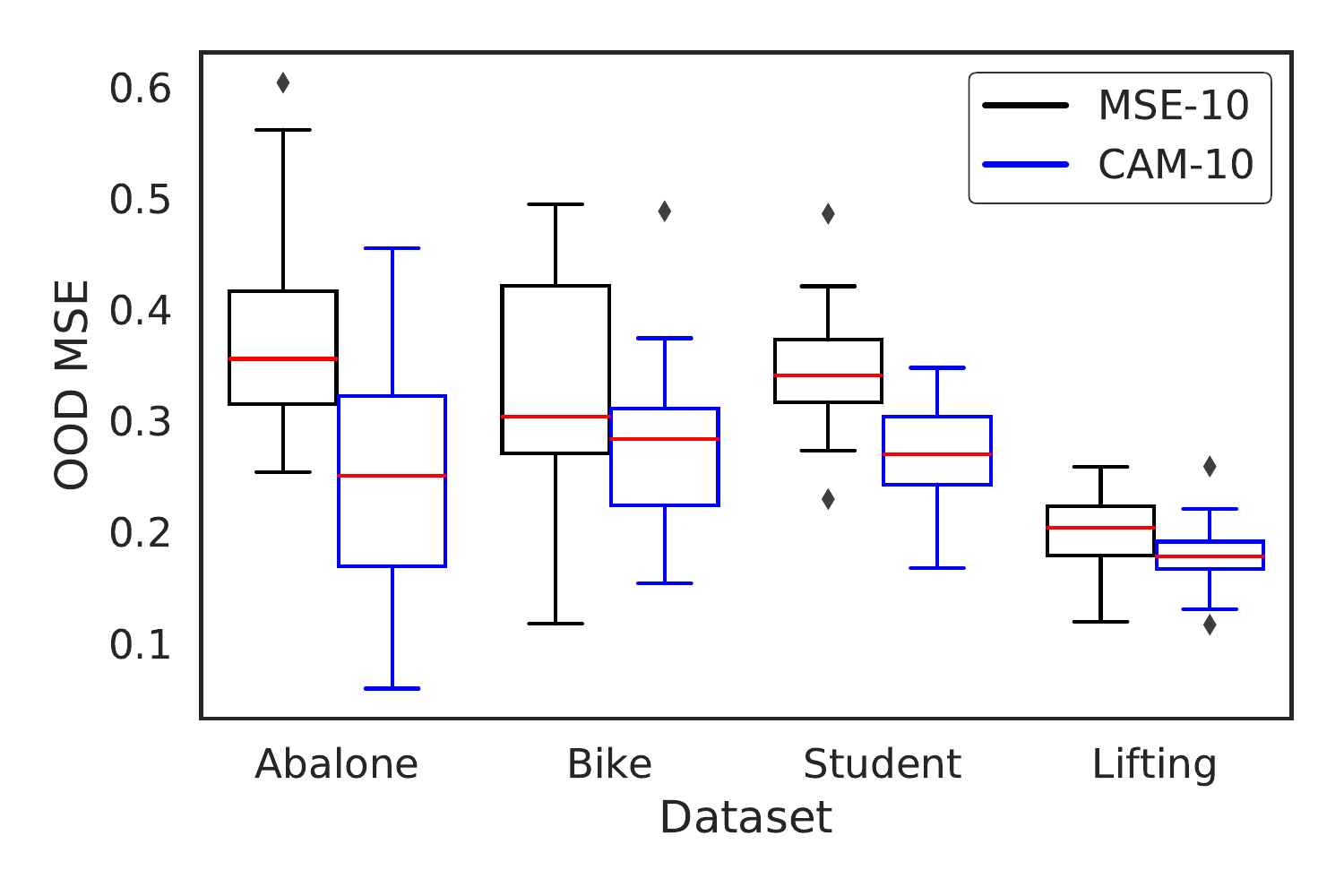}
    \caption[ ]
    {Results on real datasets: Abalone,  Bike Sharing (Bike), Student Performance on Exams (Student), and Powerlifting (Lifting). Performance of our selection method (CAM-10) in comparison to MSE-10 in terms of out-of-distribution (OOD) MSE.  Graph for inversion count in Appendix~B. For all datasets our CAM was able to select and rank the models that performed best on OOD samples.}
    \label{fig:public_results}
\end{figure}

For this experiment we focus on out-of-distribution (OOD) test performance, which is an important and practical consideration for model robustness. 
For example,  one could imagine this being applicable when data is limited and decisions are critical, such as in the healthcare industry where we require that our model generalize well for the OOD outliers that could exist at inference time.  
For each of the datasets we randomly chose a continuous variable and randomly held-out either 20\% of the lowest or greatest samples for that variable, such that these end-point values were never seen during any phase other than testing.  With the remaining 80\% we further split that into three sets of 60\%, 20\% and 20\% for model training, validation, and selection, respectively.  

For each dataset we identically trained 100 deep learning models with identical architectures on our training set.  
Each model had two hidden layers with the number of neurons equal to the number of input features.  
Both layers were initialized with Glorot uniform weights and followed by a dropout mask of 0.2 after each.  
The activation function for the hidden layers and output layer was ReLU and linear, respectively.  
Models were trained for 20 epochs using SGD with a learning rate of $10^{-4}$ with a momentum of 0.9, and stopped and saved when MSE validation loss converged and failed to improve on our validation set.  
We then evaluated and ranked each of the 100 models by their performance on our selection set, by either MSE or our CAM, and repeated this 100 times for each dataset.  
We set $\lambda$ in Eq.~\ref{eq:metric} to 1.25, which we calculated based on the results of the synthetic experiments (performance of CAM-10 over MSE-10).  
The results in terms of  model ranking and top 10\% model selection are presented in Fig.~\ref{fig:public_results} and shows that our method was able to reliably select and rank the more performant models to the OOD held-out examples for each of the datasets. 

\subsection{Transportability robustness across settings}

\begin{figure}[!t]
        \begin{subfigure}[b]{0.23\textwidth}
            \centering
            \includegraphics[width=\textwidth]{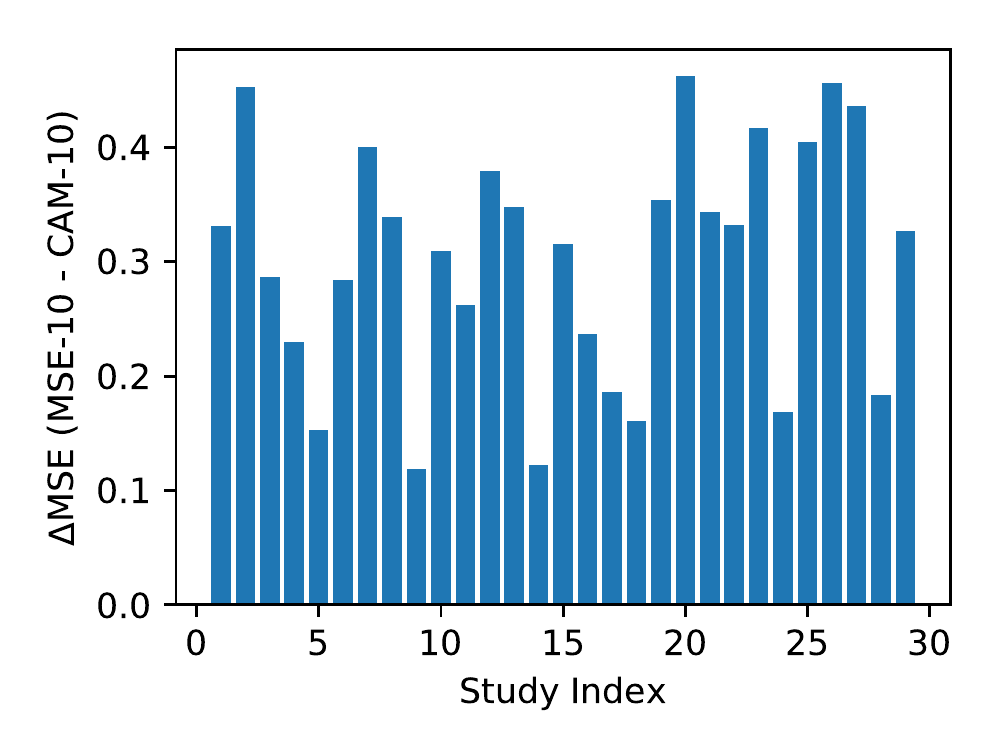}
            \caption[]%
            {Subgraph Results}    
            \label{fig:MAGGIC_MSE}
        \end{subfigure}
    \begin{subfigure}[b]{0.23\textwidth}
            \centering
            \includegraphics[width=\textwidth]{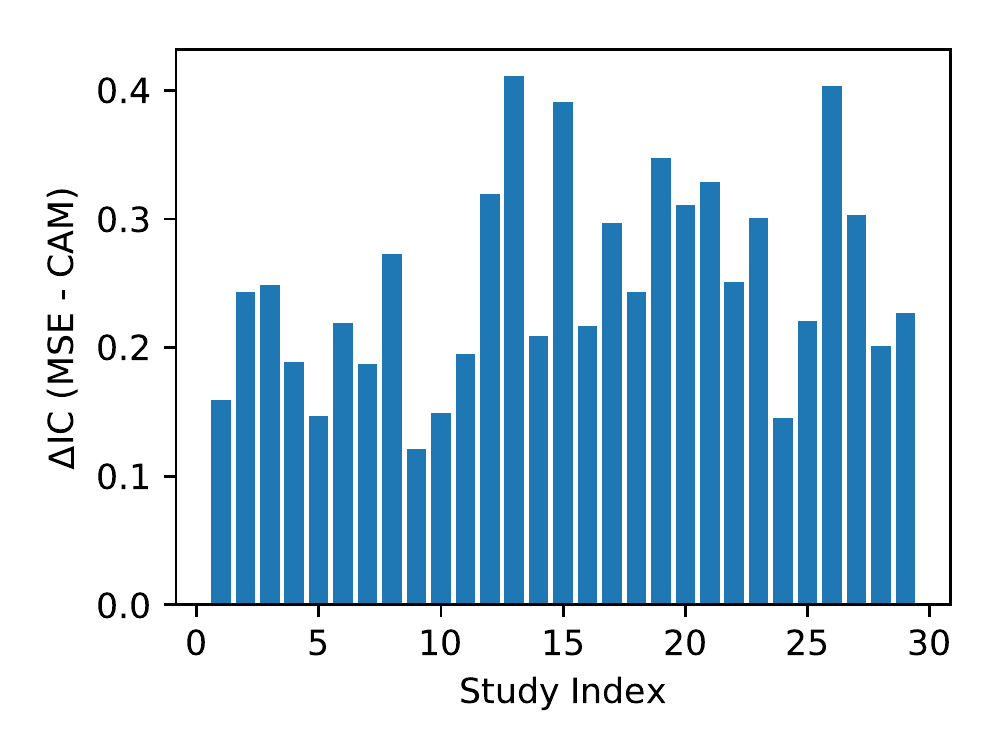}
            \caption[]%
            {Imposter Results}    
            \label{fig:MAGGIC_IC}
        \end{subfigure}
    \vspace{-10pt}
  \caption{Results on MAGGIC dataset.  (a) $\Delta$ MSE is the performance improvement using CAM over MSE in terms of top 10 model selection. (b) $\Delta$ IC is the performance improvement of using CAM over MSE in terms of inversion count.}
  \label{fig:MAGGIC}
\end{figure}

Machine learning models trained on datasets specific to one setting, such as geographic, region, or location, are susceptible to bias of these training domains. To demonstrate our method for transportability robustness across studies or populations we focused on a medical dataset collected across various geographical locations and dates. 
The Meta-analysis Global Group in Chronic heart failure database (MAGGIC) holds data for 46,817 patients gathered from 30 clinical studies or registries \cite{maggic}, and is commonly used to predict mortality in patients with heart failure.  
We used an identical experimental set up and training regime used in Section~\ref{sec:real_data}, but instead we randomly chose one of the study cohorts (with at least a thousand patients) as our training, validation and selection set, and the remainder (studies from different geographical locations) were reserved for testing.  For the target variable we predicted the patient survival time in days.  Fig.~\ref{fig:MAGGIC} shows our results in terms of $\Delta$MSE, which is the performance improvement using CAM over MSE in terms of top 10 model selection, and $\Delta$IC, which is the performance improvement of using CAM over MSE in terms of inversion count.  Additional dataset details, causal discovery steps and DAG are provided in Appendix~A.

\section{Conclusion}

We have presented a method for model selection that leverages prior knowledge in the form of a causal graph, to make predictions of the selected network more robust and invariant to unseen domains.  
We have shown promising results on synthetic data for domain shift perturbations that may occur anywhere in the DGP.  
We have shown that our method can be applied to improve existing state-of-the-art causal domain adaptation methods.  
Lastly, we have shown on four Kaggle datasets and a medical dataset how our method can be used to improve transportability to OOD examples.  
A natural next step of this research would be to investigate how our methodology can be extended into the model training phase.  
Incorporating our methodology into a trainable loss function is currently not possible due to our CAM being non-differentiable, but we hope to investigate learning techniques such as actor-critic frameworks for solving this problem.  
We hope that this work will inspire further research on the incorporation of causal knowledge into predictive models.

\bibliographystyle{icml2020}

\bibliography{causal.bib}

\appendix
\section{Public datasets}

 This section highlights the specific details for each of our public datasets. Discovered causal graphs are provided as well.

\begin{figure}[!t]
        \centering
        \begin{subfigure}[b]{0.44\linewidth}
            \centering
            \includegraphics[width=\linewidth]{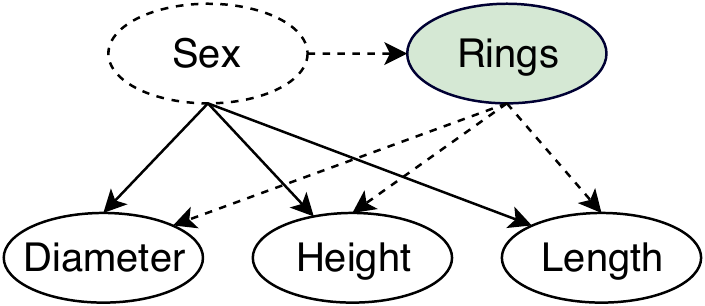}
            \caption[]%
            {Abalone dataset}    
            \label{fig:abalone}
        \end{subfigure}
        \quad
        \begin{subfigure}[b]{0.44\linewidth}  
            \centering 
            \includegraphics[width=\linewidth]{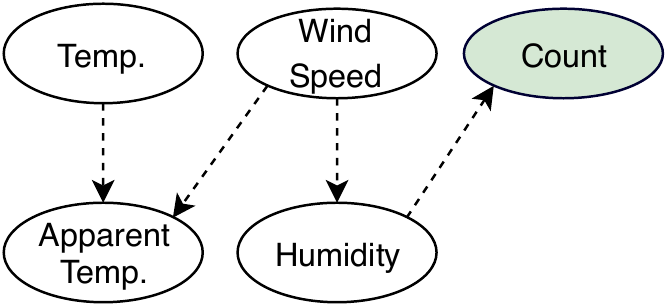}
            \caption[]%
            {Bike Sharing dataset}    
            \label{fig:bikesharing}
        \end{subfigure}
        \quad
        \centering
        \begin{subfigure}[b]{0.44\linewidth}  
            \centering 
            \includegraphics[width=\linewidth]{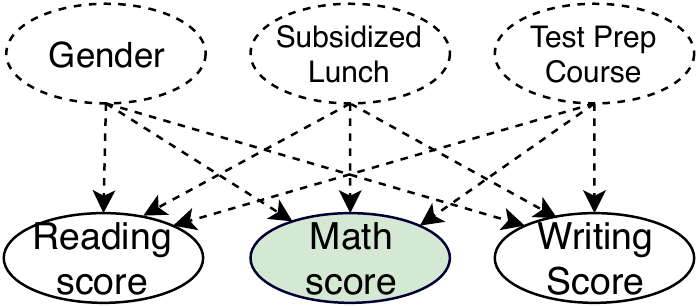}
            \caption[]%
            {Student Exams dataset}    
            \label{fig:student}
        \end{subfigure}
        \quad
    \begin{subfigure}[b]{0.44\linewidth}  
            \centering 
            \includegraphics[width=\linewidth]{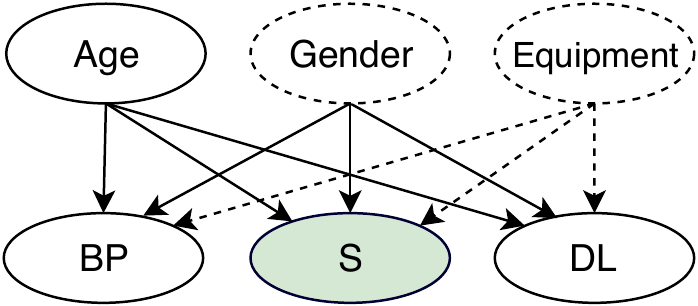}
            \caption[]%
            {Powerlifting dataset}    
            \label{fig:powerlifting}
        \end{subfigure}
    \begin{subfigure}[b]{0.77\linewidth}  
            \centering 
            \includegraphics[width=\linewidth]{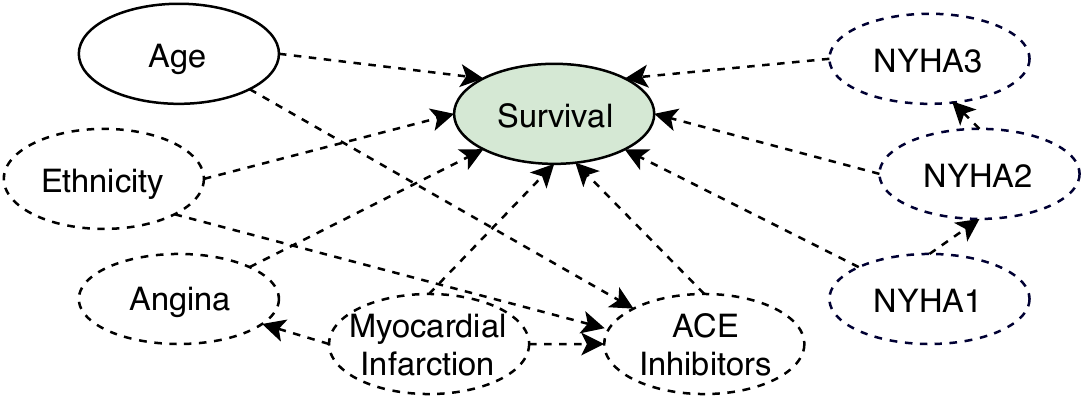}
            \caption[]%
            {MAGGIC dataset}    
            \label{fig:maggic}
        \end{subfigure}

        \caption[ ]
        {Causal DAGs used for public Kaggle dataset experiments. The green vertices represent the prediction target, and all other vertices are assumed to be input variables.  Solid bordered and dashed bordered vertices represent continuous and discrete  variables, respectively. Solid edges represent known causal connections, and dashed edges represent edges discovered using a causal discovery algorithm (GFCI).  In (d), BP, S, and DL represent bench press, squat, and deadlift, respectively.  Isolated variables that are disconnected from the DAGs because causal relationships could not be determined or discovered are not shown. In (e), for conciseness we show only the connections between causal parents of the target node (Survival).  }
        \label{fig:dags}
\end{figure}

\paragraph{Abalone dataset}  This dataset contains 4177 samples with 5 different attributes of an abalone (a type of shellfish).  We predicted the number of rings that an abalone has given its sex and shell diameter, height, and width.  It is practical to predict the number of rings of the abalone which can be used to determine the age of the mollusk. From \citet{abalone_causality}, we knew that there were direct causal links from the sex of the abalone to the shell diameter, height and length.  Additionally, using Tetrad we were able to establish the remaining causal connections (dotted edges) in Fig.~\ref{fig:abalone}.  We incorporated additional prior knowledge by restricting the discovery of any causal connections into the sex of the shellfish from any of the other variables, as this would imply they dictate the shellfish gender.

\paragraph{Bike Sharing dataset} This dataset contains 17379 hourly samples of bike rentals from 2011 to 2012 of the Capital bike share in Washington D.C.  The goal is to predict the `Count' of bike rentals based on actual weather data (`Temperature', `Humidity', `Wind Speed') and perceived temperature (`Apparent Temperature'). This dataset was previously explored for a causal application in \citet{anchor_regression}.  Although, we did not have any prior knowledge of the causal structure, we knew that the bike count could not be the cause of any of the variables, and that apparent temperature could not cause the temperature, wind speed, or humidity.  Using this prior knowledge and GFCI in Tetrad we were able to discover the DAG in Fig.~\ref{fig:bikesharing}.

\paragraph{Student Performance in Exams dataset} This dataset contains 1000 students and their standardized test performance for reading, math, and writing.  There were five other variables that were suspected have an impact on the outcome of these standardized test scores, which were the students gender, their race, their parent's level of education, if they received subsidized lunch or not, and whether or not they received a test preparation course.  When using Tetrad to discover the causal graph, we used the prior knowledge that none of the test scores could cause any of the aforementioned student attributes.  Fig.~\ref{fig:student} shows the resulting discovered graph.  The vertices for race and parental education are not shown as there was no causal relationship discovered.  However, we still included these variables in our neural network training and testing.

\paragraph{Powerlifting dataset}  This dataset is comprised of approximately 1.42 million records for the sport of powerlifting, in which competitors strive to lift the most weight according to their weight, age, and lifting equipment used.  Outcomes exist for three different lifting events: bench press, squat, and deadlift.  These events have several attempts for each lift, but in this experiment we focus on the first attempt for each as it contains the least amount of missing data.  We reduced the dataset down to only lifters that completed there first lift in all three lifting events, leaving approximately 200,000 samples.  Given that classes of powerlifting events are stratified by both age and gender, we applied ``common sense'' knowledge here assuming that both age and gender are causes for the lifting outcomes.  Not surprisingly, using Tetrad we discovered that the use of lifting equipment was also a direct cause of the lifting outcomes.  The causal graph is presented in Fig.~\ref{fig:powerlifting}.

\paragraph{MAGGIC dataset} This dataset is comprised of 46,817 patients collected across various locations to study the survival time after heart failure.  Each patient may have one or more comorbidities such as miocardial infarction or angina, which are documented in this dataset along with patient attributes such as gender or age.  We assumed as prior knowledge that the patient attributes such as gender, age, and ethnicity, could not be a descendant (effect) of any of the other observed variables.  We also assumed for prior knowledge that the last observation was the survival time (Survival in Fig.~\ref{fig:maggic}), which could not be an ancestor (cause) of any of the other observed variables. 

\paragraph{Discovered causal graphs} 
For each dataset we used either prior or ``common sense'' knowledge to assist in the discovery of the full causal graph.
If the complete causal graph wasn't known, to learn the remaining causal connections from the data we used FGES on the entire dataset using the \texttt{Tetrad} software package.  
Only the directed edges that were output in the CPDAG by FGES were considered as known edges in the causal graphs, which are shown in Fig.~\ref{fig:dags}.

%

\texttt{Tetrad} allows prior knowledge to be specified in terms of required edges that must exist, forbidden edges that will never exist, and temporal restrictions (variables that must precede other variables). Using our prior knowledge, \texttt{Tetrad} produced the causal DAGs shown in Fig.~\ref{fig:dags}, where the green vertices represent the prediction target. Solid bordered and dashed bordered vertices represent continuous and discrete variables, respectively. Solid edges represent the known causal connections, and dashed edges represent discovered edges. The \texttt{Tetrad} software package automatically handles mixed connections, i.e. edges between discrete and continuous variables.  If not using \texttt{Tetrad} for mixed variables, the method from can be used.

\section{Supplementary Experiments}

\subsection{Additional figures}

\begin{figure}[!ht]
        \centering

        \begin{subfigure}[b]{0.22\textwidth}
            \centering
            \includegraphics[width=\textwidth]{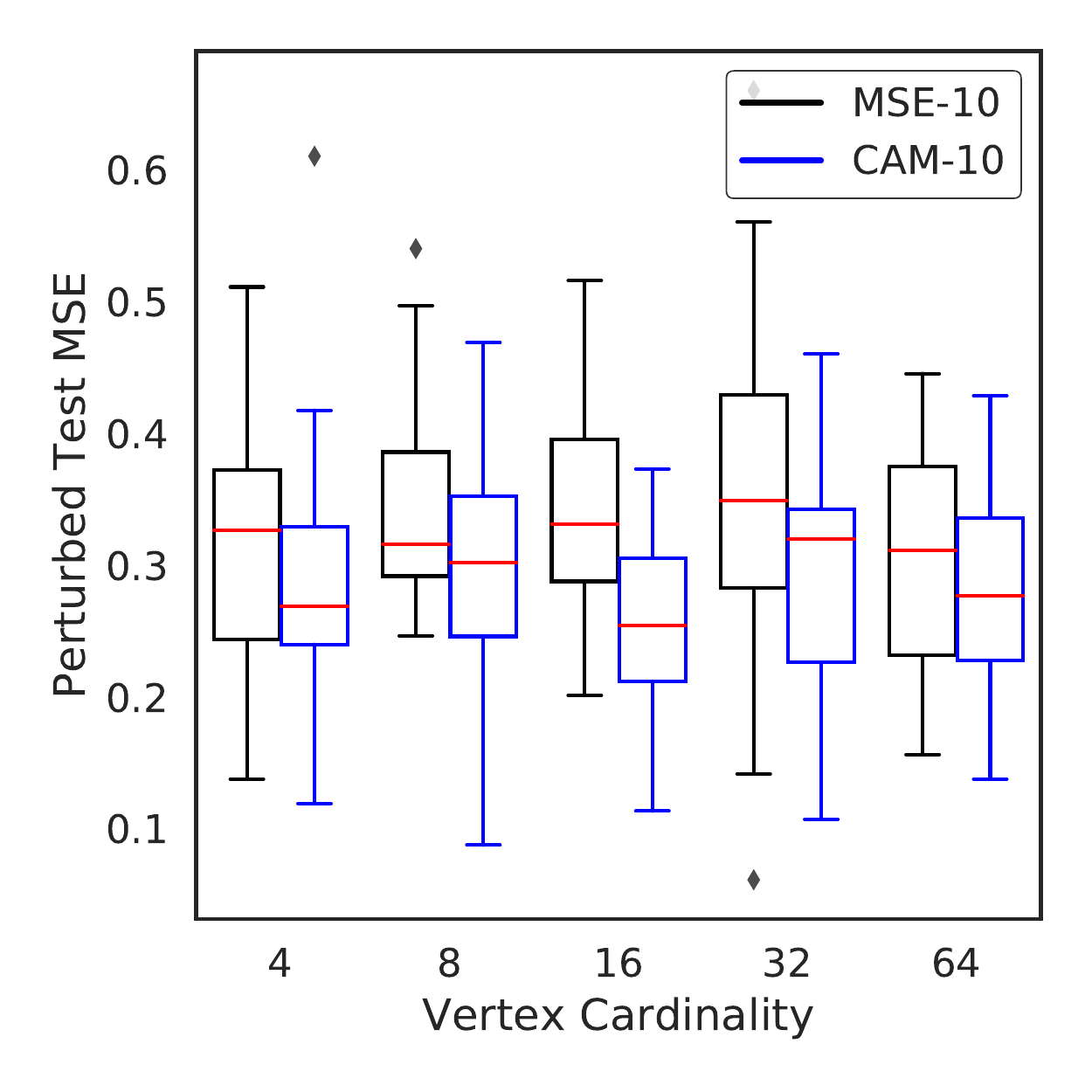}
            \caption[]%
            {MSE-10 vs CAM-10}    
            \label{fig:intervention}
        \end{subfigure}
        \centering
        \begin{subfigure}[b]{0.22\textwidth}
            \centering
            \includegraphics[width=\textwidth]{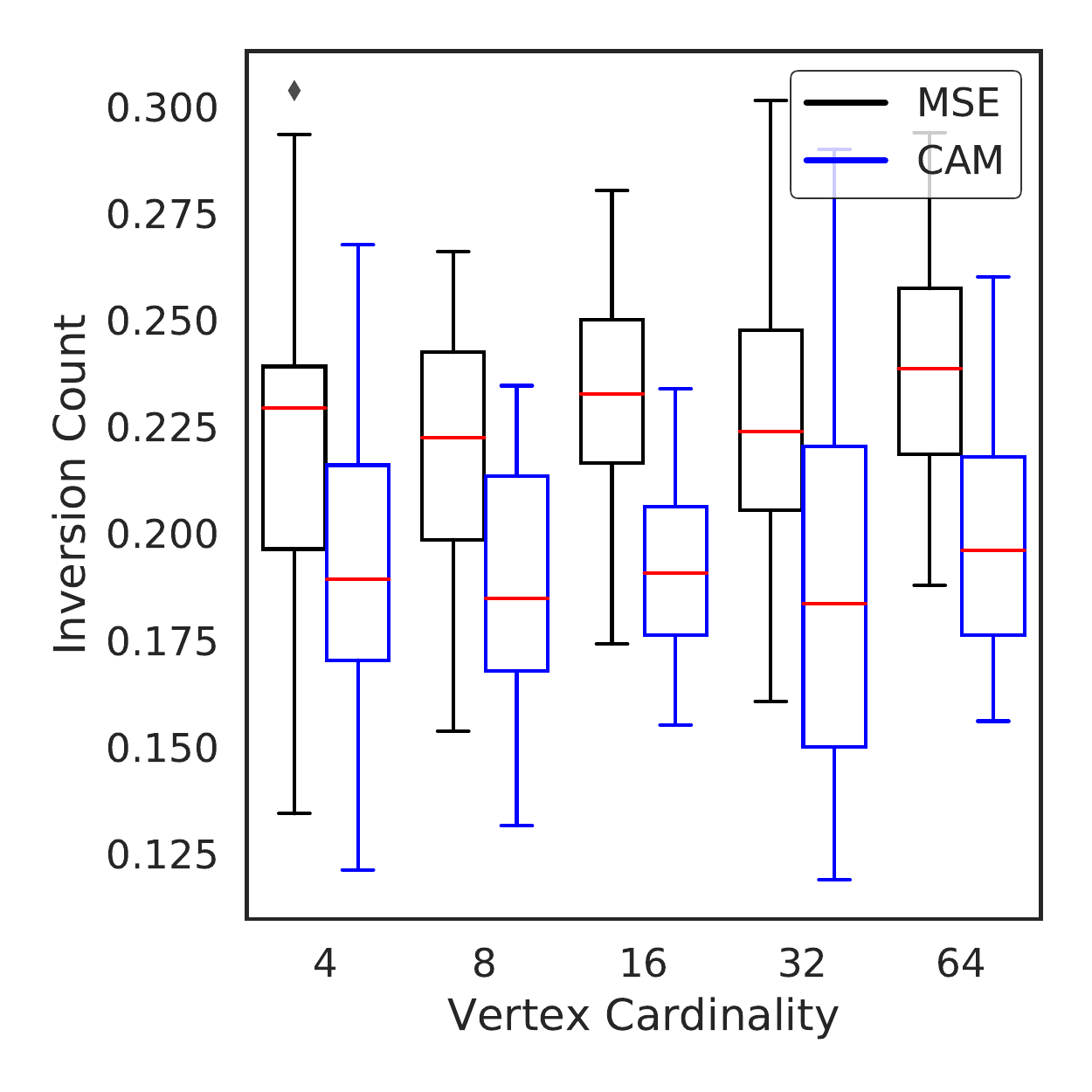}
            \caption[]%
            {Inversion count}    
            \label{fig:intervention_ic}
        \end{subfigure}
        \caption[ ]
        {Performance of proposed metric (CAM) on synthetic experiments of various vertex cardinalities. (a) We show our proposed method in terms of selecting the best models using MSE-10 and CAM-10. (b) We use an inversion count to show our proposed method in terms of ranking models by testing performance. } 
        \label{fig:synthetic_app}
    \end{figure}

\begin{figure}[!ht]
    \centering

        \includegraphics[width=0.35\textwidth]{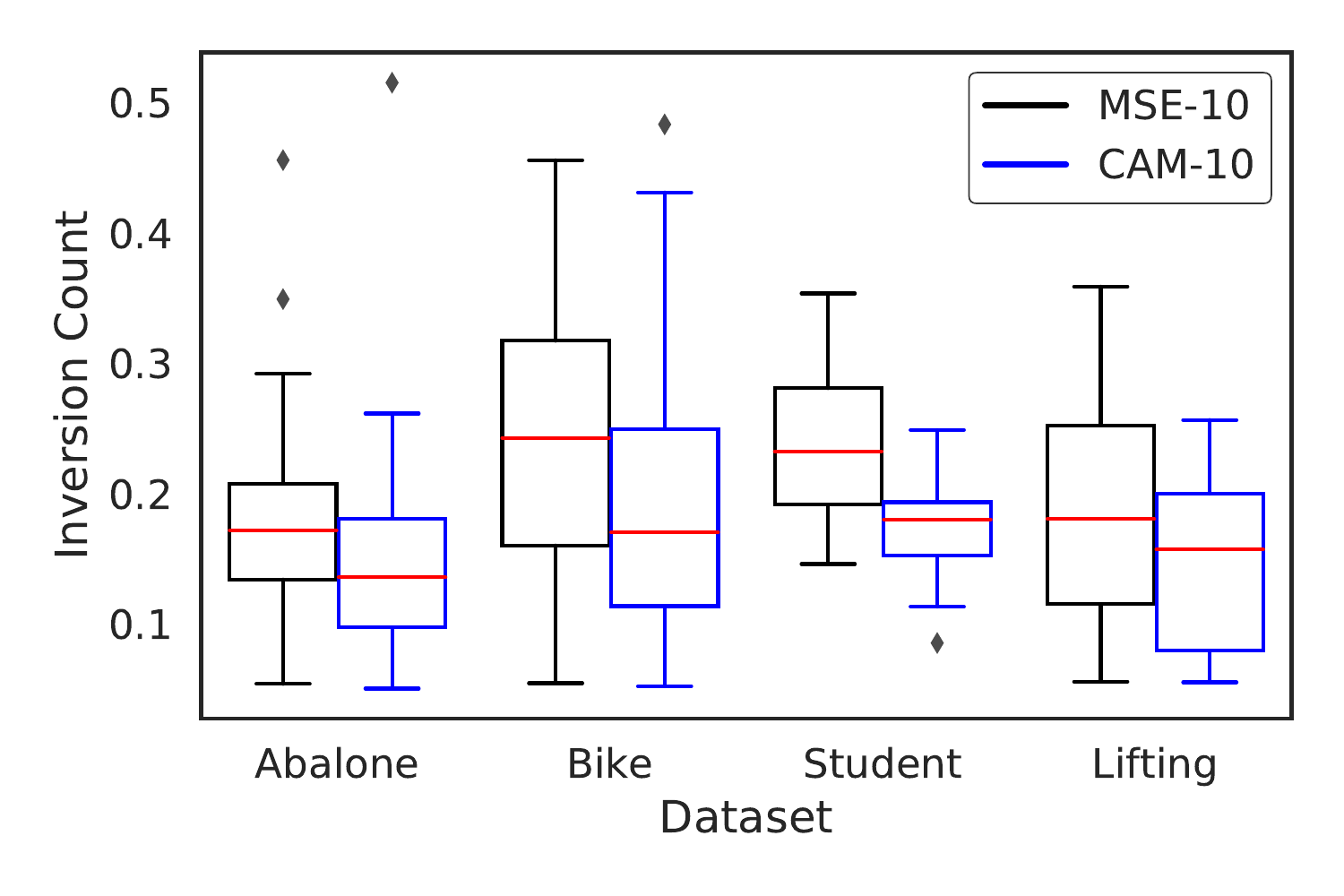}

    \caption[ ]
    {Results on real datasets: Abalone,  Bike Sharing (Bike), Student Performance on Exams (Student), and Powerlifting (Lifting). Inversion count of ranking models by MSE vs our CAM.  For all datasets our CAM was able to select and rank the models that performed best on OOD samples.}
    \label{fig:public_results_app}
\end{figure}

In this subsection we include additional figures regarding the breakdown of performance for various nodes cardinalities $n\in\{4,8,16,32,64\}$ from our synthetic experiments in Fig.~\ref{fig:synthetic_app}.  Additionally, Fig.~\ref{fig:public_results_app} shows the results of our method on real dataset in terms of inversion count.

\subsection{Other machine learning models}
\begin{figure}[!ht]
    \centering
    \begin{subfigure}[b]{0.23\textwidth}
        \centering
        \includegraphics[width=\textwidth]{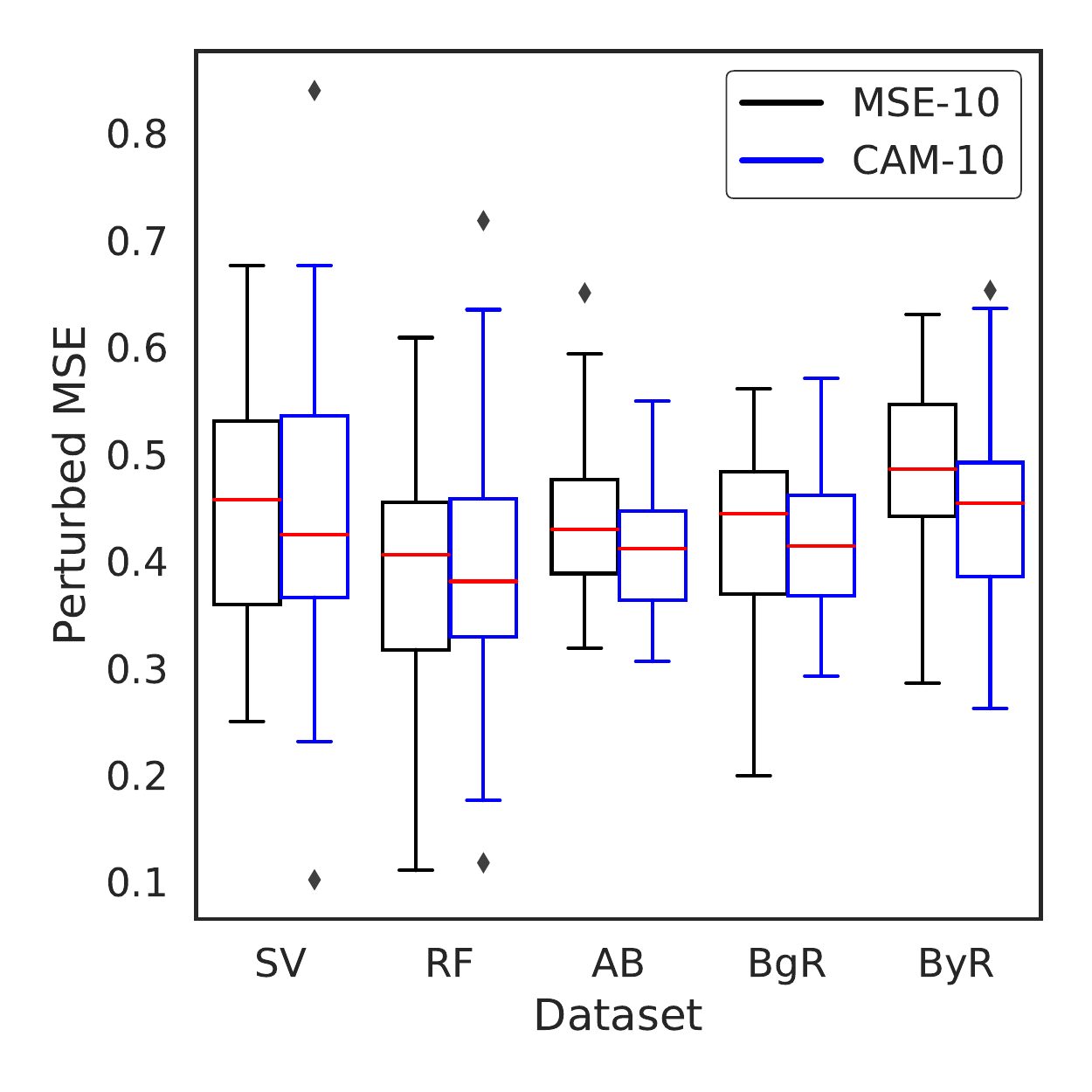}
        \caption[]%
        {MSE-10 vs CAM-10}    
    \end{subfigure}
    \centering
    \begin{subfigure}[b]{0.23\textwidth}  
        \centering 
        \includegraphics[width=\textwidth]{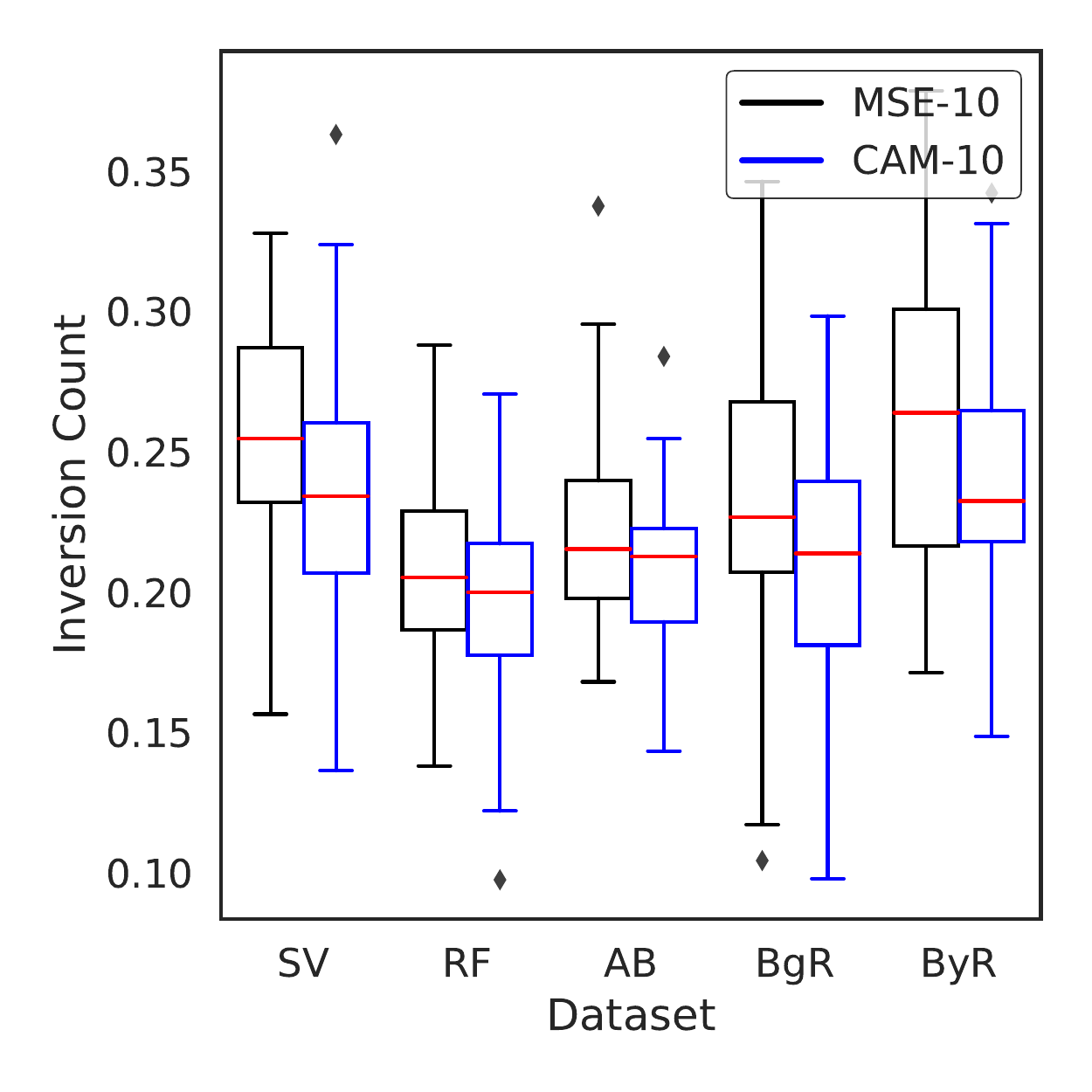}
        \caption[]%
        {Inversion count}    
    \end{subfigure}

    \caption[ ]
    {Synthetic results on other machine learning models: Support Vector (SV), Random Forest (RF), AdaBoost (AB), Bagging (BgR), and Bayesian Ridge (ByR) regressors. Results are combined over DAGs of vertex cardinalities between 4 and 64.  }
    \label{fig:others}
\end{figure}

In this section, we showcase our method being applied to methods other than neural networks.  We used the same experimental setup from the Experimental section, except we randomly generated DAGs having between 4 and 64 vertices.  We combined the results for all vertex counts as well as perturbation types (single variable and multi-variable) in Fig.~\ref{fig:others}. The results shown are the  average values in terms of MSE-10, CAM-10 and inversion count.  We show results for the following regression models: Support Vector, Random Forest, AdaBoost, Bagging Regressor, and Bayesian Ridge.  We show that our method works for more than just neural networks, and have showcased this by the results of our method on the models in Fig.~\ref{fig:others}.

\subsection{Classification}
\begin{figure}[!ht]
    \centering
    \begin{subfigure}[b]{0.18\textwidth}
        \centering
        \includegraphics[width=\textwidth]{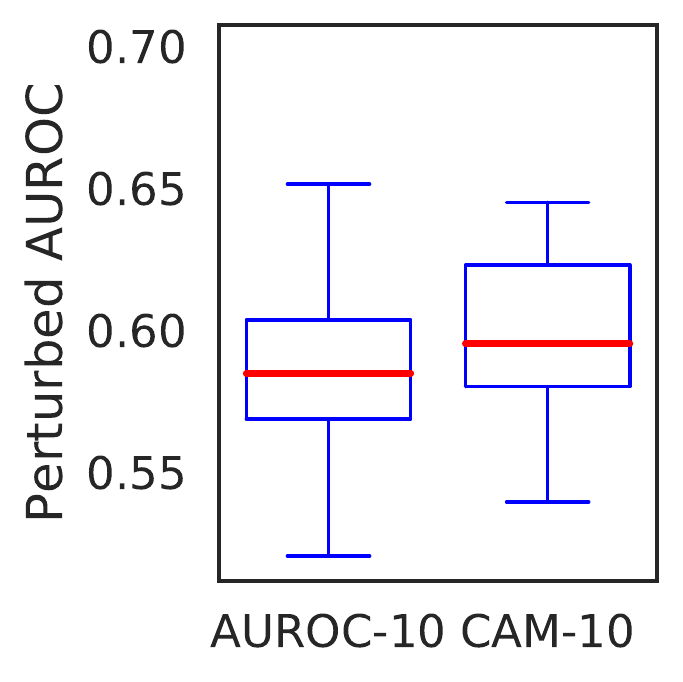}
        \caption[]%
        {AUROC}    
    \end{subfigure}
    \centering
    \begin{subfigure}[b]{0.18\textwidth}  
        \centering 
        \includegraphics[width=\textwidth]{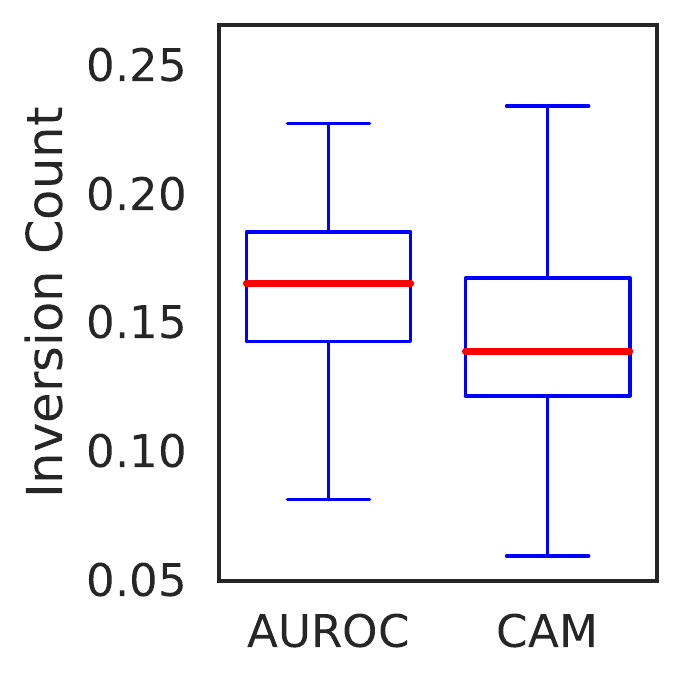}
        \caption[]%
        {Inversion count}    
    \end{subfigure}
    \caption[ ]
    {Classification experiment on synthetic data.  Presented are results in terms of selection by AUROC-10 (instead of MSE-10), which selects the top 10\% of models by AUROC.  Inversion counts comparing our ability to rank models by AUROC on the test datasets. }
    \label{fig:auroc}
\end{figure}

In this subsection we perform synthetic classification experiments rather than regression.  We use the same synthetic setup used in the Experimental section of the manuscript, but we configure our network to output a binary task.  We also randomly generate 100 DAGs having between 4 and 64 vertices.  Our target variable is drawn from a binary Bernoulli distribution with the probability being the sigmoid function applied to the target variable's value (the sum of the parent's values plus a noise term).  Again, each variable is a function of its parent's with Gaussian noise of 0 mean and 1 variance applied to each. We trained 100 identical models with a cross-entropy loss (instead of mean squared error).  Shown in Fig.~\ref{fig:auroc} are our results for model selection and ranking. For selection we use the AUROC-10 (compared to MSE-10), which selects the top 10\% of models by AUROC.  For model ranking, we use the inversion count to demonstrate our ability to rank models in terms of AUROC testing performance.

\end{document}